\documentclass[runningheads]{llncs}
\usepackage{graphicx} 
\usepackage{amsmath}
\usepackage{amsfonts}
\usepackage{mathtools}
\usepackage{url}
\renewcommand{\figurename}{Fig.}
\renewcommand{\tablename}{Table}
\def \figref #1{\figurename ~\ref{#1}}
\def \tabref #1{\tablename ~\ref{#1}}
\def \eqref #1{eq.(\ref{#1})}  
\begin{document}
\title{Identifying Critical States by the Action-Based Variance of Expected Return
}
\titlerunning{Identifying Critical States}
\author{Izumi Karino\inst{1}\orcidID{0000-0003-3267-3886} \and
Yoshiyuki Ohmura\inst{1}\orcidID{0000-0002-9158-5360} \and
Yasuo Kuniyoshi\inst{1}\orcidID{0000-0001-8443-4161}}
\authorrunning{I. Karino et al.}
\institute{The University of Tokyo, 7-3-1 Hongo, Bunkyo-ku, Tokyo 113-8656, Japan
\email{\{karino,ohmura,kuniyosh\}@isi.imi.i.u-tokyo.ac.jp}}
\maketitle              
%
%
\begin{abstract}
  The balance of exploration and exploitation plays a crucial role in accelerating reinforcement learning (RL). 
  To deploy an RL agent in human society, its explainability is also essential.
  However, basic RL approaches have difficulties in deciding when to choose exploitation as well as in extracting useful points for a brief explanation of its operation.
  One reason for the difficulties is that these approaches treat all states the same way.
  Here, we show that identifying critical states and treating them specially is commonly beneficial to both problems. 
  These critical states are the states at which the action selection changes the potential of success and failure dramatically.
  We propose to identify the critical states using the variance in the Q-function for the actions and to perform exploitation with high probability on the identified states.
  These simple methods accelerate RL in a grid world with cliffs and two baseline tasks of deep RL.
  Our results also demonstrate that the identified critical states are intuitively interpretable regarding the crucial nature of the action selection.
  Furthermore, our analysis of the relationship between the timing of the identification of especially critical states and the rapid progress of learning suggests 
  there are a few especially critical states that have important information for accelerating RL rapidly.
\end{abstract}
\section{Introduction}
Balancing exploration and exploitation is essential for accelerating reinforcement learning (RL) \cite{witten1976apparent}.
The explainability of an RL agent is also necessary when considering its application to human society.
In both of these problems, identifying a few critical states at which action selection dramatically changes the potential of success or failure would be beneficial.
These critical states are states that should be exploited and would be useful points for a brief explanation.
Previous work demonstrated that critical states are both useful for designing humanoid motion and related to human recognition \cite{kuniyoshi2004embodied}.

However, basic RL approaches do not try to detect specific states and treat all states in the same way.
The basic approach to balancing exploration and exploitation is to add a stochastic perturbation to the policy in any state in the same manner using $\epsilon$-greedy or stochastic policy, when considering application to continuous state space.
Some methods give additional rewards to rarely experienced states to bias the learning towards exploration \cite{bellemare2016unifying,pathak2017curiosity}, but they also added a stochastic perturbation in any state in the same manner.
This approach has difficulty exploring in an environment where some state-action pairs change future success and failure dramatically.
This approach makes it difficult to choose the required actions for limited transitions to successful state space.
For example, an environment might include a narrow path between cliffs through which the agent has to pass to reach a goal, or a state where the agent has to take specific action not to fall into the inescapable region due to torque limits.

Critical states contain important information that helps humans understand success and failure \cite{kuniyoshi2004embodied}.
To the best of our knowledge, no work has tried to detect such critical states autonomously.
Some studies have translated the entire policy by decomposing it into simple interpretable policies \cite{verma18a,liu2018toward}.
However, translated policies are expressed as complex combinations of simple policies and are difficult to understand.
Therefore, instead of explaining everything, extracting a few critical states that have essential information regarding success and failure is useful for creating an easily understandable explanation.

Bottleneck detection is related to critical state detection in that it captures special states.
To the best of our knowledge, a bottleneck is not defined by the action selection \cite{DBLP:conf/icml/McGovernB01,csimcsek2004using,goyal2018transfer}.
However, critical states should be defined by action selection to detect the states where specific actions are essential for success.

In this paper, we propose a method to identify critical states by the variance in the Q-function of actions and an exploration method that chooses exploitation with high probability at identified critical states.
We showed this simple method accelerates RL in a toy problem and two baseline tasks of deep RL.
We also demonstrate that the identification method extracts states that are intuitively interpretable as critical states and can be useful for a brief explanation of where is the critical point to achieve a task.
Furthermore, an analysis of the relationship between the timing of the identification of especially critical states and the rapid progress of learning suggests there are a few especially critical states that have important information that accelerates RL rapidly.
\section{Identification of critical states}
\subsection{Related work: bottleneck detection}
With respect to the identification of special states, critical state detection is related to the notion of bottleneck detection, which is used to locate subgoals in hierarchical RL.
Many studies defined bottleneck states based on state visitation frequency, that was, as ones visited frequently when connecting different state spaces \cite{csimcsek2009skill,stolle2002learning},
the border states of graph theory-based clustering \cite{csimcsek2005identifying,kazemitabar2009using}, states visited frequently in successful trajectories but not in unsuccessful ones \cite{DBLP:conf/icml/McGovernB01},
and states at which it was easy to predict the transition \cite{jayaraman2018timeagnostic}.
However, these methods can detect states that are not critical for the success of the task.
They might detect states at which the agent can return back to a successful state space by retracing its path even after taking incorrect actions.
Another method defined bottlenecks based on the difference in the optimal action at the same state for a set of tasks \cite{goyal2018transfer}.
Unfortunately, this method highly depends on the given set of tasks, cannot be applied to a single task,
and might classify almost all states as bottlenecks if the given tasks have few optimal actions in common.

\subsection{Formulation of a critical state}
Success and failure in RL are defined according to its return.
The critical states at which the action selection changes the potential for future success and failure dramatically correspond to the states where the variance in the 
return is dramatically decreased by the action selection.
We define state importance (SI) as reduction in the variance of the return caused by action selection and define critical states as states that have an SI that exceeds a threshold.
This formulation can distinguish the states that are sensitive to action selection and does not depend on a given set of tasks.
We consider the standard Markov decision process in RL. From the definition of the SI, we can calculate the SI by
\begin{align}
  \text{SI}(s) &= \text{Var}_{p(cr|s)}[C\!R] -  \mathbb{E}_{p(a|s)} [\text{Var}_{p(cr|s, a)}[C\!R]] \\
               &= \text{Var}_{\int p(a|s)p(cr|s, a)da}[C\!R]  -  \mathbb{E}_{p(a|s)} [\text{Var}_{p(cr|s, a)}[C\!R]] \\
              & \begin{multlined}[c]
                = \text{Var}_{p(a|s)}[\mathbb{E}_{p(cr|s, a)}[C\!R]] + \mathbb{E}_{p(a|s)} [\text{Var}_{p(cr|s, a)}[C\!R]]\\
                - \mathbb{E}_{p(a|s)} [\text{Var}_{p(cr|s, a)}[C\!R]]
                \end{multlined} \\
               &= \text{Var}_{p(a|s)}[\mathbb{E}_{p(cr|s, a)}[C\!R]],
\end{align}
where $cr$ denotes the return as an abbreviation of cumulative reward, $s$ and $a$ denote the state and action, respectively.
We considered that the agent selected action $a$ based on a distribution $p(a|s)$ at state $s$ and selected the actions to take at the following states based on the current policy $\pi$.
With this setup and the following definition of the Q-function:
\begin{align}
  \text{SI}(s) = \text{Var}_{p(a|s)}[Q^\pi(s, a)]\label{eq:state_importance}.
\end{align}
We used a uniform distribution as $p(a|s)$ to consider the changes in return of all actions at the current state.
We can calculate the SI without estimating the complex distributions of return $p(cr|s)$ and $p(cr|s, a)$.

We regard the states whose SI is greater than a threshold as critical states.
We set this threshold by assuming critical states make up $100q\%$ of all states.
To calculate the threshold, we first calculate the SIs of all states, if possible. If the state space is too large to calculate the SIs of all states, we calculate the SIs of recently visited states instead.
Then, we set the threshold that separates the upper $100q\%$ of SIs from the lower SIs.
We used a ratio of critical states of $q=0.1$ in all the subsequent experiments.

\subsection{Exploration with exploitation on critical states}
We propose an exploration method that chooses exploitation with a high probability of $k$ on the identified critical states and chooses an action based on any exploration method in other situations.
This method increases exploration in the areas where the agent can transit with only a limited set of actions and decreases the likelihood of less promising exploration.
We choose exploration at the identified critical state with a small probability because the estimated exploitation during learning could still be wrong.
We used an exploitation probability of $k=0.95$ in all the experiments.
\section{Experiments}
\subsection{Cliff Maze Environment}\label{subsec:toy}
\begin{figure}[t]
  \centering
  \includegraphics[width=0.7\textwidth]{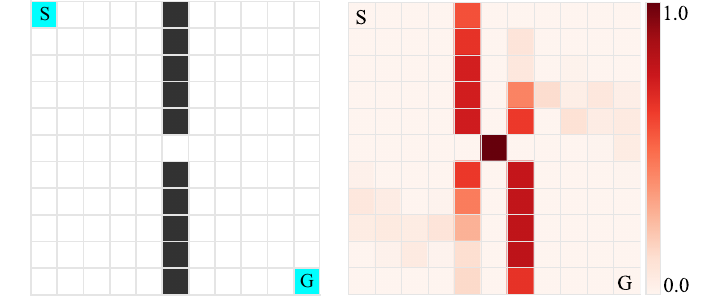}
  \caption[Cliff maze and SIs.]{Cliff maze and SIs. 
  Left: The cliff maze environment has cliffs in the center column in all cells except for the center cell. 
  The initial state is the upper-left corner cell, and the goal state is the lower-right corner cell. 
  The agent receives a reward of +1 when it reaches the goal state and a reward of -1 when it reaches a cliff state. 
  An episode terminates when the agent visits the goal state or a cliff state. 
  Right: SIs of each cell in the cliff maze. The values are normalized to [0, 1].
  }\label{fig:toy}
\end{figure}

We employed the cliff maze environment, shown on the left of \figref{fig:toy}, as the test environment. This environment includes a clear critical state at which action selection divides future success and failure dramatically.
A cliff is located in all cells of the center column of the maze except at the center cell.
The states are the 121 discrete addresses on the map. 
The initial state is the upper left corner, and the goal state is the lower right corner.
The actions consist of four discrete actions: up, down, left, and right.
The agent receives a reward of +1 at the goal state and -1 at the cliff states.
An episode terminates when the agent visits either a cliff or the goal.
The state between the two cliffs should be the most critical state.
We evaluated the proposed identification method of the critical states and the exploration performance, which
was evaluated by the number of steps required for the agent to learn the optimal shortest path.

The agent learned the Q-function via Q-learning \cite{watkins1992q}.
The Q-function was updated every episode sequentially from the end of trajectory with a learning rate of $0.3$. 
We evaluated the exploration performance of the proposed method by comparing the proposed method (based on $\epsilon$-greedy) with the simple $\epsilon$-greedy method.
The proposed method chooses exploitation if the current state is a critical state with high probability of $k$ and chooses a random action with low probability 1 - $k$ depending on the current state.
If the current state is not a critical state, this method selects action as $\epsilon$-greedy does with exploration ratio $\epsilon'$.
We controlled the exploration ratio $\epsilon'$ using the following equation  so that the exploitation probabilities of both the methods were as similar as possible.
\begin{align}
  \epsilon' &= \cfrac{\epsilon - (1-k)q}{1-q}. \label{eq:calc_eps}
\end{align}
This control means that exploitation at a critical state is essential rather than randomly taking exploitation with certain probability regardless of the state.
The exploration ratio $\epsilon$ of $\epsilon$-greedy was annealed linearly from 0.905 at the initial step to 0.005 at the final 100k-th step.
The exploration ratio $\epsilon'$ of the proposed method was correspondingly decayed from 1.0 to 0.0.
We conducted 100 experiments each with both methods with different random seeds.

The right part of the \figref{fig:toy} shows the SI calculated after a 100k-step exploration.
The SI was maximum at the state between the two cliffs. 
Intuitively, this state is a critical state, where the action selection divides the potential for success and failure dramatically.
\tabref{tab:toy} shows the mean and standard deviation of the steps required to learn the shortest path.
The proposed method reduced the median number of required steps by about 71\% and the mean by about 68\% compared to the simple $\epsilon$-greedy method.
To analyze the results, we used the Wilcoxon rank-sum test, and the $p$ value was $5 \times 10^{-13}$.
Its result shows that the proposed method significantly reduces the number of exploration steps required to learn the optimal shortest path in the cliff maze.
\begin{table}[htbp]
  \centering
  \caption{Number of exploration steps required to learn the optimal shortest path.}\label{tab:toy}
  \begin{tabular}{c|c|c|c|c}
    Method             & ~$p$ value ($N=100$)~  & ~median~ & ~mean~ & ~std~  \\ \hline 
    $\epsilon$-greedy  &   -                  & ~22815~  & ~26789~ & ~19907~  \\ \hline
    ~Proposed~           & $5 \times 10^{-13}$  & 6450     & 8468    & 6443   \\ \hline
  \end{tabular}
\end{table}
\vspace{-5mm}    
%
\subsection{Scalability to large state-action space}
We evaluated the proposed method on two baseline tasks of deep RL to confirm whether it could scale up to high-dimensional complex tasks.
The statistical significance of the mean of the acquired return was calculated with a Bayes regression of the time-series model, as defined in the Appendix.
We show the results for a ratio of critical states $q=0.1$ below, since changing the ratio $q$ from 0.05 to 0.5 did not make a significant difference in the results.
\subsubsection{Large discrete space.}\label{subsubsec:large_descrete}
We used the Arcade Learning Environment Atari2600 Breakout \cite{bellemare2013arcade} as an example of a high-dimensional discrete space task.
The agent learned the Q-function with the deep Q-network (DQN) \cite{mnih2015human}.
The comparison procedure was the same as that in the experiment of \ref{subsec:toy} except that we linearly annealed $\epsilon$ from 0.905 at the initial step to 0.005 at the final 2,000k-th step.
We calculated the threshold of the SI every 1,000 steps with the 1,000 most recently visited states because it is difficult to calculate the SI of all the states in the large state space.
We evaluated the return after every 1,000 steps of exploration by choosing exploitation from the initial state to the end of the episode.

The hyperparameter settings were as follows.
The environment setting was the same as that of \cite{mnih2015human} except that we did not use flickering reduction and we normalized each pixel to the range [0, 1].
We approximated Q-function with a convolutional neural network (CNN).
The CNN had three convolutional layers with size (32$\times$32, 8, 4), (64$\times$64, 4, 2), and (64$\times$64, 3, 1) with rectified linear unit (ReLU) activation.
The elements in the brackets correspond to (filter size, channel, stride).
A fully-connected layer of unit size 256 with ReLU activation followed the last convolutional layer.
A linear output layer followed this layer with a single output for each action. 
We used the target network \cite{mnih2015human} as the prior work.
The target network was updated every 500 steps, and the Q-function was updated every four steps after 1,000 steps.
The size of the replay buffer was $10^4$, the discount factor was 0.99, and the batch size was 64.
We used the Adam optimizer \cite{DBLP:journals/corr/KingmaB14} and clipped the gradient when the L2 norm of the gradient of the DQN loss function exceeded 10.
We initialized the weights using Xavier initialization \cite{pmlr-v9-glorot10a} and the biases to zero.

\begin{figure}[t]
  \centering
  \includegraphics[width=0.8\textwidth]{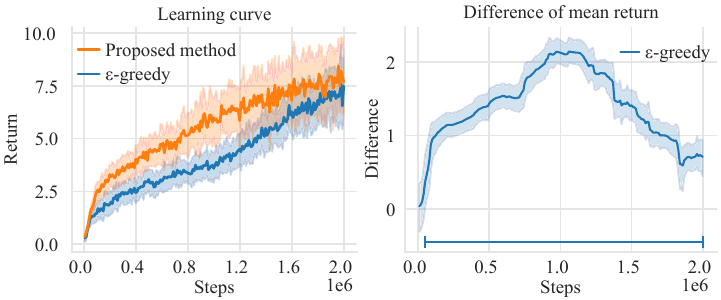}
  \caption[Learning curve in the Breakout task]{Learning curve in the Breakout task.
  Returns are the average of the ten most recent evaluations to smooth the curve for visibility.
  Left: Each colored line and shading corresponds to the mean and standard deviation of the average return at each evaluation, respectively.
  Right: The line and shading correspond to mean and 95\% credibility interval of the estimated difference of the return from the proposed method at each step, respectively. 
  The horizontal line at the bottom indicates the interval over which the 95\% credibility interval is greater than 0. The sample size is 25 for each method.
  }\label{fig:dqn_result}
\end{figure}
\begin{figure}[t]
  \centering
  \includegraphics[width=0.8\textwidth]{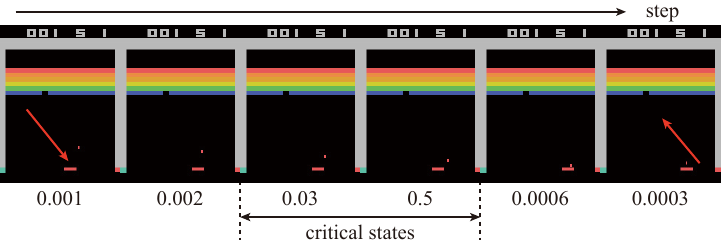}
  \caption[Critical states identified in the Breakout task]{Critical states identified in the Breakout task. The numbers correspond to the SIs. 
  The arrow indicates the direction the ball is moving to.
  }\label{fig:dqn_visualize}
\end{figure}
The left part of \figref{fig:dqn_result} shows the change in the acquired return during learning.
The right part of \figref{fig:dqn_result} shows the mean and the 95\% credibility interval of the difference of mean return between the proposed method and $\epsilon$-greedy method.
The difference in the mean return was greater than 0 from the 50k-th step to the final 2000k-th step.
A statistical difference existed in the latter 97.5\% steps, at least in the sense of Bayes regression under our modeling (detailed modeling is shown in the Appendix).
This result shows that the proposed method acquired a higher return in fewer exploration steps in the latter 97.5\% of the learning process than $\epsilon$-greedy.
\figref{fig:dqn_visualize} shows part of a fully exploited trajectory and the identified critical states.
The agent could identify critical states at which the action selection is critical.
At these critical states, the agent must move the plate toward the ball. Otherwise, the ball will fall.
The fifth frame is not a critical state because it has already been determined that the agent can hit the ball back without depending on action selection.
This enables the agent to explore where to hit the ball.
%
\subsubsection{Continuous space.}\label{subsubsec:continuous}
We tested our method in an environment with continuous state space and continuous action space.
We used the Walker2d environment of the OpenAI Gym \cite{openaigym} baseline tasks.
The RL algorithm used to train the Q-function was a soft actor-critic \cite{pmlr-v80-haarnoja18b} without entropy bonus.
The coefficient of the entropy bonus was set to 0 to evaluate how our method works on a standard RL objective function.
We approximated the SI with the Monte Carlo method by sampling 1,000 actions from a uniform distribution.
The threshold of the SI was calculated in the manner used in the experiment in large discrete space.
The proposed method chooses exploitation with high probability of $k$ and chooses action based on a stochastic policy in other situations.
We compared the exploration performance of the proposed method to two methods. 
One of the methods is called Default, and chooses action based on a stochastic policy at all states.
The other one is EExploitation, which chooses exploitation with probability $e$ and uses a stochastic policy with a low probability of $1-e$.
We controlled the exploitation ratio $e$ so that exploitation probabilities of both the proposed method and EExploitation were as similar as possible.
This control means that exploitation at a critical state is essential rather than randomly taking exploitation with certain probability regardless of the state.
The exploitation ratio $e$ was set to $e=qk$.
We evaluated the return every 1,000 steps of exploration by choosing exploitation from the initial state to the end of the episode.
 
The hyperparameter settings were as follows.
We used neural networks to approximate the Q-function, V-function, and policy.
The Q-function and V-function and the policy had two fully-connected layers of unit size 100 with ReLU activations.
The output layers of the Q-function and V-function networks were linear layers that output a scalar value.
The policy was modeled as a Gaussian mixture model (GMM) with four components.
The output layer of the policy network outputted the four means, log standard deviations, and log mixture weights.
The log standard deviations and the log mixture weights were clipped in the ranges of [-5, 2] and [-10,], respectively.
To limit the action in its range, a sampled action from the GMM passed through a tanh function and then was multiplied by its range.
L2 regularization was applied to the policy parameters.
We initialized all the weights of neural networks using Xavier initialization \cite{pmlr-v9-glorot10a} and all the biases to zero.
The size of the replay buffer was $10^6$, the discount factor was 0.99, and the batch size was 128. We used the Adam optimizer \cite{DBLP:journals/corr/KingmaB14}.
We updated the networks every 1,000 steps.

\begin{figure}[t]
  \centering
  \includegraphics[width=0.8\textwidth]{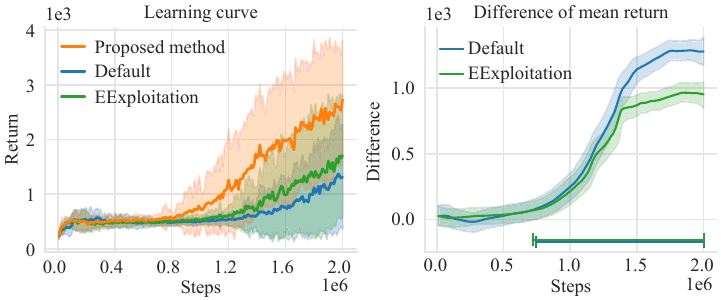}
  \caption[Learning curve in the Walker2d task.]{Learning curve in the Walker2d task.
  Returns were averaged over the ten most recent evaluations to smooth the curve for visibility.
  Left: The colored line and shading correspond to the mean and standard deviation of the averaged return at each evaluation, respectively.
  Right: The lines and shading correspond to the means and 95\% credibility intervals of the estimated difference of the return from the proposed method at each step, respectively. 
  Each horizontal line at the bottom indicates the interval where 95\% interval is greater than zero. The sample size is 25 for each method.
  }\label{fig:sac_result}
\end{figure}
\begin{figure}[t]
  \centering
  \includegraphics[width=0.8\textwidth]{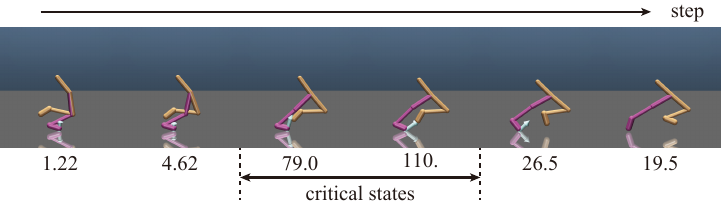}
  \caption[Critical states identified in the Walker2d task]{Critical states identified in the Walker2d task. The numbers correspond to the SIs. 
  The arrows indicate contact force.
  }\label{fig:sac_visualize}
\end{figure}
The left part of \figref{fig:sac_result} shows the trend of the return acquired through learning.
The right part of \figref{fig:sac_result} shows the mean and the 95\% credibility interval of the difference in the mean returns of the proposed method and each of the other methods.
The difference of the mean return was greater than 0 from the 720k-th step to the final 2000k-th step for Default, and from the 740k-th step to the final 2000k-th step for EExploitation.
A statistical difference existed in the latter 63\% steps, at least in the sense of Bayes regression under the modeling (the detailed model is given in Appendix).
This result shows that the proposed method acquired a higher return with fewer exploration steps in the latter 63\% of the learning process.
\figref{fig:sac_visualize} shows a part of a fully exploited trajectory and the identified critical states.
The agent could identify critical states at which the action selection was critical.
The agent had to lift the body by pushing the ground with one leg while it lifted and moved forward with the other leg.
Otherwise it would stumble and fall.
%
\subsection{Relationship between the identification of critical states and the speed of learning}\label{subsec:knack}
\begin{figure}[t]
  \centering
  \includegraphics[width=0.9\textwidth]{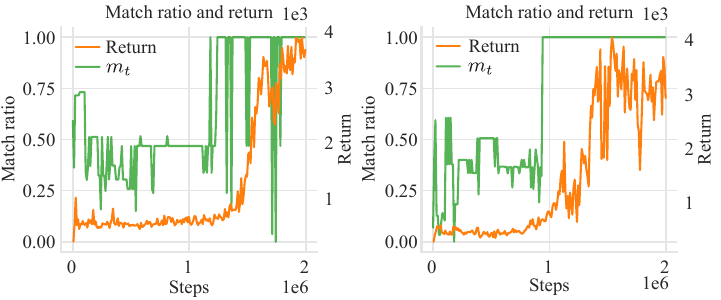}
  \caption{Two examples of the match ratio. Here, $m_t$ corresponds to the maximum match ratio of the top-ten critical states between the $t$-th step and the final step.
  }\label{fig:match_ratio}
\end{figure}
We analyzed the relationship between the timing of the identification of the especially critical states and the rapid progress of learning in the Walker2d task.
The analysis process was as follows.
The parameters of the Q-function at each $ t $ step and all states visited by the agent over the entire training were saved.
We calculated the SI of all of the states with the saved Q-function's parameters.
We regarded states whose SI values were in the top ten as especially critical states.
We calculated the minimum Euclidean distance $d_t$ between the top-ten critical states of the final step and those of the $t$-th step, assuming that the SIs at the final step were correct.
We calculated the $d_t$ from the 1k-th step to the 2000k-th step in 10k step intervals.
We calculated the match ratio $m_t$ at each step $t$ as $m_t = 1.0 - d_t / d_{\text{max}}$, where $d_{\text{max}}$ is the maximum value in $d_t$.
The value of match ratio $m_t$ becomes closer to 1 as at least one of the top-ten critical states of $t$ step become closer to those of the final step.

\figref{fig:match_ratio} shows two examples of the relationship between the timing of the identification of the top-ten critical states and the progress of learning.
The return curves rapidly rose shortly after the agent begins to identify the top-ten critical states.
This phenomenon might be interpreted as the agents "got the knack of" the task.
This tendency was observed in some but not all agents.
This result suggests there are a few critical states that had important information for accelerating RL rapidly.
\section{Conclusion}
In this paper, we assumed that the critical states at which the action selection dramatically changes the potential for future success and failure are useful
for both exploring environments that include such states and extracting useful information for explaining the RL agent's operation.
We calculated the critical states using the variance of the Q-function for the actions.
We proposed to choose exploitation with high probability at the critical states for efficient exploration.
The results show that the proposed method increases the performance of exploration and the identified states are intuitively critical for success.
An analysis of the relationship between the timing of the identification of especially critical states and the rapid progress of learning suggests there are a few especially critical states that have important information for accelerating RL rapidly.

One limitation of our proposed method is that it does not increase the performance of exploration in environments with no critical states, where it does not decrease the performance significantly. 
Exploitation at falsely detected critical states is not critical for future success in such environments. 
In addition, this approach does not greatly increase the possibility of being trapped in local optima because it does not aim to reach critical states but just chooses exploitation when the visited state is a critical state. 
We decided the ratio of critical states $q$ to use to set the threshold and the exploitation ratio $k$ in advance. 
These values could be decided automatically during learning using the uncertainty of the Q-function with a Bayesian method. 
The proposed identification method extracted useful contents for a brief explanation, but their interpretation was constructed by a human. 
A method to construct a brief explanation using the identified critical states is future work. 
\subsubsection{Acknowledgments.}
This paper is based partly on results obtained from a project
commissioned by
the New Energy and Industrial Technology Development
Organization (NEDO), and partly on results obtained from research
activity in  Chair for Frontier AI Education, School of Information
Science and Technology, The University of Tokyo.
The final authenticated publication is available online at \url{https://doi.org/10.1007/978-3-030-61609-0_29}.
\appendix
\section*{Appendix}
\subsubsection{Calculation of the statistical difference of two learning dynamics.}
A statistical test to confirm the difference between algorithms is necessary for deep RL,
 because large fluctuations in performance have been observed even as a result of different random seeds \cite{henderson2018deep}.
We cannot evaluate the exploration steps required to learn the optimal solution because the optimal solution is difficult to define in a large state space.
A statistical test has been performed on the difference of the mean of return at a certain step \cite{henderson2018deep},
 but this makes determining the evaluation points arbitrary.
In addition, independently treating each timestep does not properly handle the data because the return curve is time-series data.
In this paper, we used Bayesian regression for the return curves by modeling them as random walks depending on one previous evaluation step.
We consider there to be a statistical difference when the 95\% credibility interval of the time series of the average difference of returns between the proposed method and each method exceeds 0.
We model the return curve by
\begin{align}
  \mu[t] &\sim \mathcal{N}(\mu[t-1], \sigma_\mu) \label{eq:stat-1}\\
  \delta_X[t] &\sim Cauchy(\delta_X[t-1], \sigma_X) \label{eq:stat-2}\\
  R_{\text{Proposed}}[t] &\sim \mathcal{N}(\mu[t], \sigma_R) \label{eq:stat-3}\\
  R_X[t] &\sim \mathcal{N}(\mu[t] - \delta_X[t], \sigma_R) \label{eq:stat-4}.
\end{align}
We suppose that return of the proposed method $R_{\text{Proposed}}[t]$ is observed from the true value $\mu[t]$ with observation noise $\sigma_R$.
This observation noise is caused by evaluating only several episodes instead of evaluating all possible trajectories.
We assume that the return of a comparison method $R_X [t]$ has fluctuation $ \delta_X[t] $ from $ \mu[t] $, and that observation noise $\sigma_R$ has been added.
Here, $\mu[t]$ is modeled as a random walk with a normal distribution depending on the previous evaluation step with a variance of $ \sigma_\mu $.
In addition, $\delta_X[t]$ is modeled as a random walk with a Cauchy distribution depending on the previous evaluation step with a parameter of $ \sigma_\mu $ so that large fluctuations can be handled robustly.
The statistical difference $\delta_X[t]$ was calculated by estimating the posterior with a Markov chain Monte Carlo (MCMC) method.
We modeled the priors of all variables as a no-information uniform prior and initialized the first MCMC sample of $\mu[t]$ by the mean return of the proposed method at each evaluation step $t$.
\bibliographystyle{bib/splncs04}
\bibliography{bib/RL}
\end{document}